\documentclass{article}
\usepackage{ijcai26}
 
\usepackage{times}
\usepackage{soul}
\usepackage{url}
\usepackage[hidelinks]{hyperref}
\usepackage[utf8]{inputenc}
\usepackage[small]{caption}
\usepackage{graphicx}
\usepackage{amsthm}
\usepackage{booktabs}
\usepackage{multirow}
\theoremstyle{definition}
\newtheorem{definition}{Definition}
\usepackage{algorithm}
\usepackage{algorithmic}
\title{The Annotation Scarcity Paradox in Low-Resource NLP Evaluation:\\
A Decade of Acceleration and Emerging Constraints}

\author{
    Vukosi Marivate$^{1,2}$
    \affiliations
    $^1$Data Science for Social Impact (DSFSI) \& African Institute for Data Science and Artificial Intelligence (AfriDSAI), University of Pretoria \\
    $^2$Lelapa AI
    \emails
    vukosi.marivate@cs.up.ac.za
}
\begin{document}
 
\maketitle
 
\begin{abstract}

Over the past decade, low-resource natural language processing (NLP) has experienced explosive growth, propelled by cross-lingual transfer, massively multilingual models, and the rapid proliferation of benchmarks. Yet this apparent progress masks a critical, insufficiently examined tension: the deep sociolinguistic expertise required to evaluate increasingly complex generative systems is severely strained, inequitably distributed, and structurally marginalised. We present a critical narrative survey of low-resource NLP evaluation (2014--present), tracing its evolution across three phases: early heuristic optimism, the illusions of top-down benchmark scaling, and the current era of generative bottlenecks. We conceptualise the \emph{Annotation Scarcity Paradox}, the structural friction arising when the technical capacity to scale models vastly outpaces the sovereign human infrastructure required to authentically evaluate them. By examining extractive data pipelines, undercompensated ``ghost work'', and language data flaring, we argue that this paradox threatens the epistemic validity of reported progress. We survey emerging responses---including data augmentation, model-based evaluation, participatory curation, and annotation-efficient approaches via item response theory and active learning---and assess their equity and validity trade-offs. We close with a practitioner call to action, arguing that overcoming this bottleneck requires a paradigm shift from transactional data extraction to relational, community-embedded evaluation rooted in epistemic governance, data sovereignty, and shared ownership.
\end{abstract}
 
\section{Introduction}
 
The past decade has produced an extraordinary proliferation of work in low-resource Natural Language Processing (NLP)~\cite{joshi2020state,alabi2025charting,belay2025rise}. Shared tasks, multilingual benchmarks, and community-driven resource creation efforts have dramatically expanded the set of languages for which computational tools exist. Projects such as MasakhaNER~\cite{adelani2022masakhaner}, AfriSenti~\cite{muhammad-etal-2023-afrisenti}, AmericasNLP Shared Tasks~\cite{mager-etal-2021-findings} and SEACrowd~\cite{lovenia-etal-2024-seacrowd}, exemplify the ambition and reach of this movement.
 
However, this rapid accumulation of benchmarks and reported performance gains has masked a structural fragility: the evaluation pipelines underpinning this progress depend critically on human annotators, linguists, and community members whose capacity is finite, whose labour is often uncompensated, and whose involvement is frequently shallow. As the demand for evaluation data has scaled, the human infrastructure required to produce high-quality annotations has not kept pace. Compounding this, the purported gains of the Large Language Model (LLM) era are still accumulating disproportionately to high-resource languages~\cite{blasi2022systematic,ahuja2023mega,adelani2025irokobench,ojo2025afrobench}, widening the gap between what is technically possible and what is practically evaluable for the majority of the world's languages. On one of the continent's best language coverage benchmark, AfroBench, the strongest proprietary model (GPT-4o) achieves an average of only 59\% across 64 African languages~\cite{ojo2025afrobench}---a figure that itself reflects a curated, relatively well-resourced subset of Africa's 2,123 languages~\cite{adebara2025ai}.

We term this the \textbf{Annotation Scarcity Paradox}, defined as follows:

\begin{definition}[Annotation Scarcity Paradox]
The structural friction arising—most directly evidenced in the African NLP context, and, we argue, in analogous forms across other under-resourced language communities—when the technical capacity to produce and scale NLP models outpaces the human infrastructure, encompassing annotator availability, deep linguistic expertise, community participation, and epistemic governance (whose knowledge matters), required to authentically evaluate them. The paradox is not merely logistical but structural, shaping what can be known and what counts as progress in low-resource NLP.
\end{definition}

This survey makes the following contributions:

\begin{itemize}
    \item We argue that the trajectory of low-resource NLP evaluation across three chronological phases (2014--2018, 2019--2022, 2023--present) has produced a \emph{structural}, not merely logistical, bottleneck in the human capacity required for credible evaluation.
    \item We introduce and operationalise the \emph{Annotation Scarcity Paradox} as a conceptual framework for understanding how extractive data pipelines, undercompensated annotation labour, and data sovereignty deficits jointly undermine the epistemic validity of reported progress.
    \item We identify concrete directions for transitioning from transactional data extraction to relational, community-embedded evaluation, including pluralistic evaluation frameworks, transparent annotation reporting norms, and community-owned data infrastructure.
\end{itemize}

\medskip\noindent\textbf{Methodological note.}
This survey proceeds as a critical narrative review~\cite{snyder2019literature,grant2009typology}, drawing on a selection of representative works rather than an exhaustive corpus search. Papers were drawn primarily from proceedings at ACL, ACM venues, and associated community workshops (AfricaNLP, AmericasNLP, WiNLP), as well as shared task system description papers. Selection was guided by three thematic criteria: papers documenting evaluation methodology or benchmark construction for low-resource languages; papers examining annotation practices, community engagement, or data governance; and papers that illuminate the structural dynamics of the field rather than solely reporting model performance. Phase boundaries reflect recognised inflection points: 2019 marks the emergence of massively multilingual models (mBERT~\cite{devlin2019bert}, XLM-R~\cite{conneau2020unsupervised}) and large-scale cross-lingual benchmarks (XTREME)~\cite{hu2020xtreme}; 2023 marks the dominance of generative evaluation paradigms and the rise of LLM-as-a-judge methods.
The three-phase periodisation proposed here is corroborated by the bibliometric
analysis of \citeauthor{belay2025rise}~\shortcite{belay2025rise}, which tracks 2.2K
AfricaNLP papers from 2005--2025 and documents publication growth and topic shift
patterns consistent with the phase boundaries identified here.
The author has been an active participant in the low-resource NLP
ecosystem described here, particularly within African language NLP communities; this positionality is both the analytical warrant for the claims advanced and a transparency obligation to the reader.

This paper is also situated within the tradition of Critical Data Studies
(CDS)~\cite{iliadis2016critical,couldry2019costs}, which examines how data collection,
curation, and deployment practices reproduce and entrench existing power asymmetries.
Central CDS concepts---data colonialism~\cite{couldry2019costs}, epistemic
injustice~\cite{dencik2019exploring}, and the racialisation of algorithmic
systems~\cite{benjamin2019race}---provide the vocabulary through which the Annotation
Scarcity Paradox is here conceptualised. Where CDS has largely engaged with data as a
social and political phenomenon at the level of platforms and governance, this paper
extends those concerns into the technical specifics of NLP evaluation infrastructure,
asking what CDS implies for benchmark design, annotation labour, and the measurement
of model progress in low-resource language communities.

\section{The Early Boom (2014--2018): Initial Efforts and Optimism}
 
The period from 2014 to 2018 was characterised by pioneering efforts to extend NLP tools beyond the handful of high-resource languages that had historically dominated the field. Early work focused on establishing practical pipelines for resource creation and developing fundamental processing tools for previously underserved languages \cite{king2015practical}. During this time, researchers across the Global South actively began constructing localised benchmarks to ensure non-Western languages were represented in the growing data ecosystem. In the African context, localised institutional efforts laid crucial groundwork; for instance, \citeauthor{eiselen2014developing} \shortcite{eiselen2014developing} developed foundational text corpora and core processing technologies for ten South African languages. Parallel momentum was visible in Southeast Asia through multinational collaborations like the Asian Language Treebank \cite{thu2016introducing}, and in India, where researchers built massive, open-source datasets such as the IIT Bombay English-Hindi Parallel Corpus \cite{kunchukuttan2018iit}.

Crucially, this period culminated in the realisation that technical algorithms alone could not solve the resource gap~\cite{nekoto2020participatory,hershcovich2022challenges}.
The technical barriers encountered by researchers, combined with the friction of historically extractive data collection practices, highlighted an urgent need for data sovereignty and epistemic governance. This catalysed the rise of grassroots movements working towards increasing the participation of indigenous languages in the epistemic discourse of NLP. We saw the founding of the Masakhane Research Foundation—driven directly by the collaborative efforts of African researchers—and Widening NLP (WiNLP) at ACL, building on traditions of grassroots movements within the wider AI space. 
 
\subsection{Resource Creation and Early Benchmarks}
 
The 2014--2018 era saw researchers begin developing corpora and annotation frameworks for
African, Asian, and Indigenous languages, frequently working with small
teams of linguists and community volunteers. The excitement of this
period was genuine: even modest datasets enabled meaningful downstream
experiments, and cross-lingual transfer from high-resource languages
offered a seemingly scalable path to broader coverage.

A primary technical driver of progress during this era was cross-lingual transfer, which sought to leverage data-rich languages to bridge the data gap in low-resource settings by mapping representations across languages \cite{adams2017cross}. Concurrently, the transition to Neural Machine Translation (NMT) sparked new methodological approaches, with researchers exploring universal models capable of translating low-resource languages by sharing syntactic and lexical representations across multiple languages \cite{gu2018universal}. These technical advancements were soon paired with localized evaluations addressing specific linguistic families. \citeauthor{abbott2018towards} \shortcite{abbott2018towards} laid early groundwork for NMT applied to African languages, highlighting both the severe data sparsity challenges and the potential for neural architectures to overcome them. Similarly, \citeauthor{mager2018challenges} \shortcite{mager2018challenges} mapped the unique morphological and infrastructural challenges facing indigenous languages of the Americas, signalling a growing need for tailored, region-specific methodologies.
 
\subsection{Optimism and Its Limits}
 
The optimism of this phase rested on an implicit assumption: that cross-lingual transfer and bootstrapped resources could substitute for deep, language-specific human investment \cite{joshi2020state}. This assumption was rarely examined critically. While zero-shot architectures and universal models demonstrated mathematical ingenuity, they frequently reduced complex morphological and syntactic phenomena into universalist frameworks that inherently favoured high-resource pivot languages, typically English \cite{mager2018challenges}. 

Consequently, efforts to build corpora were often opportunistic rather than systematic. The field relied heavily on scraping readily available, but narrowly focused texts (such as religious translations \cite{agic2019jw300} 
or government proceedings), resulting in severe domain mismatch and a lack 
of true sociolinguistic representation. This era was largely defined by a 
top-down, extractive approach to natural language processing. Languages were 
frequently treated as mere data points to solve a technical optimization 
challenge, disconnected from the communities who actually spoke them 
\cite{bird2020decolonising}. 

Because of this disconnect, there was an absence of epistemic
governance within the research lifecycle. The researchers building the models
rarely possessed lived experience with the languages, and the communities
generating the data had no sovereignty over how their linguistic heritage
was utilised, licensed, or deployed. Annotation teams were small, sometimes
consisting of a single annotator per language, and inter-annotator agreement
was inconsistently reported \cite{mager2018challenges}. 

The foundations for rigorous evaluation were present in principle but
seldom fully realised in practice. Because native speakers were largely
excluded from the development loop, evaluations relied heavily on automated
metrics applied to noisy, out-of-domain test sets, creating a false sense
of empirical progress~\cite{birhane2022values}. 
Ultimately, the technical limitations of this phase
made it increasingly evident that algorithmic advancements could not outpace
the fundamental need for community-driven stewardship. The friction generated
by these historically extractive practices highlighted the necessity for a
radical realignment in how NLP research was conducted, creating the exact
vacuum that community-led grassroots movements would soon rise to fill.

\section{The Scaling Challenge (2019--2022): Benchmarking and the Illusion of Progress}
 
Between 2019 and 2022, the field entered a phase of rapid benchmark
proliferation, driven largely by the advent of Massively Multilingual 
Language Models (MMLMs) such as mBERT and XLM-R. Shared tasks multiplied, 
multilingual leaderboards emerged, and performance on standardised 
benchmarks became the primary currency of scientific contribution. However, 
this scaling often masked profound underlying deficiencies in how low-resource 
languages were processed.
 
\subsection{Proliferation of Benchmarks and Shared Tasks}
 
The number of shared tasks targeting low-resource languages grew
substantially during this period. Workshops associated with major venues
(ACL, EMNLP, COLING) hosted annual competitions that attracted
large numbers of submissions and generated significant visibility for
low-resource NLP research. This era saw a divergence in benchmarking 
approaches. On the one hand, massive, top-down evaluation suites like 
XTREME \cite{hu2020xtreme} attempted to create universal metrics for 
cross-lingual generalization. On the other hand, grassroots initiatives 
focused on building authentic datasets from the ground up, evidenced by the 
AI4D Language Program challenges for dataset creation 
\cite{siminyu2020ai4d,siminyu2021ai4d,orlic2021outreach}, which actively 
funded and structured community-driven resource generation.
 
\subsection{Pressure for Performance and Comparability}
 
The competitive dynamics of shared tasks and overarching leaderboards created 
intense incentives for optimising benchmark performance, frequently at the 
expense of deeper sociolinguistic understanding. \citeauthor{rodriguez-etal-2021-evaluation}~\shortcite{rodriguez-etal-2021-evaluation}
and \citeauthor{joshi2020state}~\shortcite{joshi2020state} raised crucial concerns about benchmark bias and
the degree to which reported gains reflected genuine linguistic
competence rather than dataset-specific artefacts or overfitting. The 
standardisation of evaluation formats, while enabling broad comparability, 
flattened important morphological and syntactic distinctions between language 
families. Furthermore, models scaled to accommodate over a hundred languages 
often suffered from capacity dilution; researchers found that state-of-the-art 
MMLMs systematically failed to represent the nuances of low-resource languages 
unless subjected to intensive, language-specific adaptive fine-tuning 
\cite{alabi2022adapting}.
 
\subsection{Emerging Awareness of Limitations and the Participatory Shift}
 
By the end of this phase, critical voices within the community had begun
to rigorously question the assumptions underlying rapid benchmark scaling. 
Concerns centred on the shallow involvement of speaker communities in benchmark
design and the concentration of annotation labour among a small pool of
multilingual academics. It became evident that separating the technical creation 
of benchmarks from the communities who speak the languages perpetuated the 
extractive practices of the previous decade. This realisation birthed the 
participatory research paradigm, championed by movements like Masakhane Research Foundation\footnote{\url{https://www.masakhane.io/}} (and others such as AmericasNLP\footnote{\url{https://www.americasnlp.org}}, ), which 
demonstrated that integrating native speakers into every node of the NLP 
development pipeline was not merely an ethical imperative, but a technical 
necessity for building robust models \cite{nekoto2020participatory}. These 
concerns and subsequent methodological shifts foreshadowed the severe data and 
compute bottlenecks that would become increasingly visible in the era of Large 
Language Models (LLMs).
 
\section{The Annotation Scarcity Paradox (2023--present): Limits and New Directions}
From 2023 onwards, the Annotation Scarcity Paradox has become an increasingly explicit concern in the literature and in community discussions. As the field transitioned from discriminative or highly constrained tasks to open-ended  generative AI, the sheer volume of data required to align and evaluate models met the hard reality of limited human capital. Several converging pressures have brought this tension to the fore.
 
\subsection{The Rising Cost of High-Quality Evaluation}
 
As LLMs have become central to NLP research, the complexity and time required for meaningful human evaluation has increased substantially~\cite{bender2021dangers}. Unlike earlier eras where automated metrics like BLEU or F1 could provide a rough proxy for performance, evaluating the outputs of generative models (for fluency, cultural appropriateness, safety, and factual accuracy) requires evaluators with deep linguistic and cultural competence, not simply literacy in the target language~\cite{manduchi2024challenges}. For many low-resource languages, such evaluators are exceptionally scarce. Consequently, there is a growing realisation that model capabilities are severely bottlenecked not just by training data, but by the availability of qualified humans to verify the outputs~\cite{kholodna2024llms,chiu2025culturalbench}.
 
\subsection{Challenges of Community Engagement and Data Sovereignty}
 
Meaningful participation from speaker communities has proven difficult to sustain at scale. The power dynamics between resource-rich research institutions (often located in the Global North) and language communities have received growing critical attention \cite{nekoto2020participatory}. Historically, data pipelines have been largely extractive: language is  frequently treated as a raw resource to be harvested rather than a living  system governed by its speakers. Consequently, communities are often asked to provide intensive annotation labour without commensurate recognition, financial benefit, or agency over the resulting models. 

This friction has catalysed a strong push for \textit{data sovereignty}~\cite{effoduh2026decolonizing,birhane2020algorithmic} and  equitable data licensing frameworks (such as Kaitiakitanga~\cite{taiuru2021kaitiakitanga}, CARE Principles~\cite{carroll2023care}, Esethu Framework~\cite{rajab2025esethu}, Nwulite Obodo Open Data License (NOODL) \cite{okorie2025s,okorie2025addressing}), advocating that communities must retain  ownership over their linguistic heritage and dictate how it is deployed.
However, implementing these equitable governance structures requires  significant time and trust-building, further constraining the speed at which ``scale'' can be achieved. Similarly, the pool of computational linguists with  deep expertise in specific low-resource languages remains limited, creating structural constraints on the quality of evaluation that can be produced.

\subsection{The Hidden Labour and Ethical Debt of Scale}

The structural constraints of scaling generative models are not merely technical; they are deeply tied to extractive labour practices and poor resource management. As \citeauthor{birhane2021large}~\shortcite{birhane2021large} extensively critique, the prevailing ``bigger is better'' paradigm inherently relies on the indiscriminate scraping of massive datasets, which disproportionately encodes and amplifies structural harms against marginalised groups. The sheer scale of these datasets
creates severe barriers to responsible data filtering and ethical auditing, frequently shifting the burden of safety onto the very communities most likely to be harmed by the resulting models (see, on a smaller scale, the work in \citeauthor{abdulmumin-etal-2024-correcting}~\shortcite{abdulmumin-etal-2024-correcting}, which corrects an African-focused MT dataset that is widely used in the field).

This brings to light the precarious reality of data workers in the Global South. \citeauthor{okolo2024reforming}~\shortcite{okolo2024reforming} has documented how the safety and alignment of global AI systems, such as toxicity filtering and reinforcement learning from human feedback (RLHF), rely heavily on under-compensated ``ghost workers'' in regions like Africa. While resource-rich institutions reap the economic and technological benefits of these models, the psychological toll and manual labour of data annotation are outsourced. This dynamic exacerbates the global AI divide, prompting scholars to advocate for comprehensive, localised data governance frameworks that protect data workers and enforce equitable regulation across the continent. 

\begin{figure}[t]
    \centering
    \includegraphics[width=0.90\columnwidth]{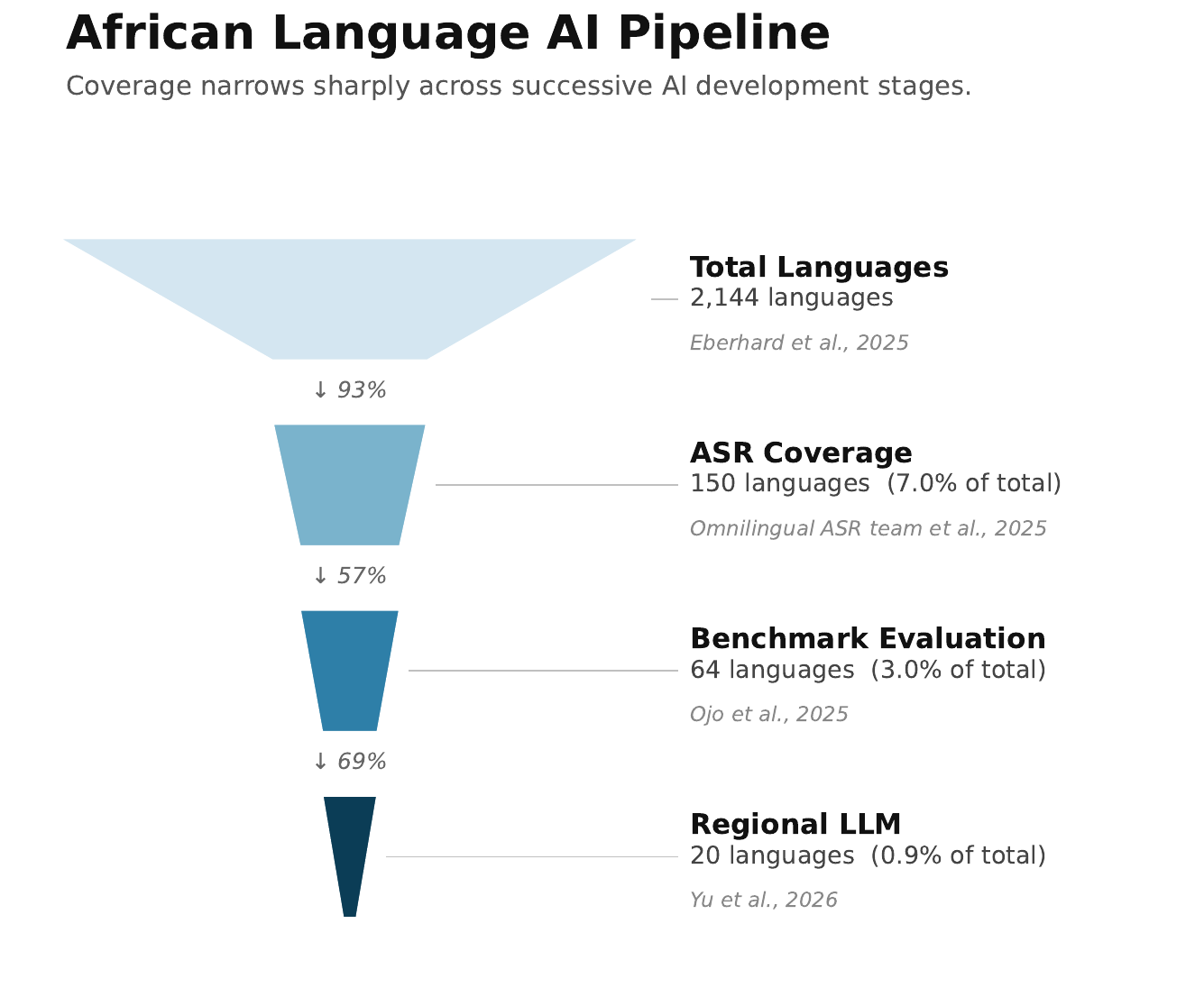}
    \caption{The African language AI pipeline bottleneck. Each bar shows
    independent coverage counts at successive stages of the AI pipeline:
    2,144 total languages~\protect\cite{eberhard2025ethnologue}; at least
    150 with any ASR system~\protect\cite{meta2025omnilingual} (estimated
    lower bound); 64 covered by a major NLP
    benchmark~\protect\cite{ojo2025afrobench}; 20 served by a regional
    LLM~\protect\cite{yu2026afriquellm}. Languages at each stage do not
    necessarily overlap with adjacent stages --- the figure illustrates the
    stark disparity in scale across the pipeline, not a strict attrition
    sequence.}
    \label{fig:africa-funnel}
\end{figure}

Compounding this exploitative paradigm is the phenomenon recently termed ``language data flaring'' by \citeauthor{adebara2025ai}~\shortcite{adebara2025ai}. Paralleling the wasteful burning of natural gas during oil extraction, language data flaring captures the systemic neglect, poor digitisation practices, and extreme under-utilisation of existing African linguistic resources. While high-resource languages are aggressively harvested to push the boundaries of LLM capabilities, vast amounts of low-resource data remain siloed in physical archives or inaccessible digital formats. This wasteful management means that local communities are simultaneously exploited for their annotation labour while being starved of the digital infrastructure necessary to bring their own native languages into the epistemic discourse of modern NLP. The coverage gap is stark: across Africa, at least 150 of 2,144 languages~\cite{eberhard2025ethnologue} have any ASR coverage~\cite{meta2025omnilingual}, fewer than 80 appear in any NLP benchmark~\cite{adebara2025ai,ojo2025afrobench,adelani2025irokobench,pazabench2026}, and only 20 are served by a regional LLM (Figure~\ref{fig:africa-funnel}); in Southeast Asia, SEACrowd covers 36 of an estimated 1,300 indigenous languages~\cite{lovenia-etal-2024-seacrowd}; in South Asia, IndicGenBench spans 29 of roughly 800 languages~\cite{l2024indicgenbench}; and across the Americas, indigenous NLP benchmarks collectively cover fewer than 25 of over 1,000 indigenous languages~\cite{mager-etal-2021-findings,ebrahimi2022americasnli}.

\subsection{Emerging Alternative Approaches}

Facing this paradox, researchers have begun exploring approaches that reduce, redistribute, or democratise the demands on human evaluators. These include:

\begin{itemize}
\item \textbf{Augmentation:} A complementary strategy is to maximise the value of already-annotated data through augmentation. Widely applied in computer vision to improve model robustness, data augmentation artificially expands available training and evaluation sets. For low-resource NLP, this offers a partial workaround for annotation scarcity~\cite{csahin2022augment}, though the choice of technique matters: different augmentation methods affect morphological and syntactic properties differently~\cite{feng2021survey,dhole2023nl,chen2023empirical}, and naive application risks amplifying existing biases or obscuring the linguistic phenomena under study.
    \item \textbf{Model-Based Evaluation:} Using advanced LLMs as proxies for
    human raters (``LLM-as-a-judge'') to scale evaluation dynamically
    \cite{zheng2023judging}. However, this approach carries significant risk
    in low-resource settings, as the ``judge'' models themselves suffer from
    severe pre-training data imbalances and frequently fail to capture
    localised cultural nuances.
    \item \textbf{Global Participatory Curation:} Massive, globally distributed annotation efforts designed specifically to bridge the instruction-tuning gap. A canonical example is the Aya initiative \cite{singh2024aya}, which  involved collaborators from over 100 countries in a human-curated, participatory framework to build instruction-following datasets for 65 languages.  For Automated Speech Recognition datasets we have the Mozilla Common Voice\footnote{\url{https://commonvoice.mozilla.org/en}}~\cite{ardila2020common} project which worked with volunteers to collect scripted speech.
    However, even participatory campaigns at scale carry inherent tensions: volunteer fatigue and dropout across long-running efforts can produce uneven language coverage~\cite{klie-etal-2023-lessons}; quality control becomes more difficult as annotator linguistic backgrounds vary widely; and "global" participation risks over-sampling diaspora or digitally-connected speakers relative to in-country communities most directly affected by the resulting models. The source of data used in such campaigns may still face challenges, such as where Mozilla Common Voice gets its source textual data~\cite{de2022localising}. 
    \item \textbf{Annotation-Efficient Evaluation via IRT and Active Learning:}
Rather than collecting human judgements uniformly across all evaluation items, Item Response Theory (IRT) models the \emph{difficulty} and \emph{discriminative power} of individual test examples with respect to model capability \cite{lalor-etal-2019-learning}. Applied to LM evaluation, IRT identifies the small subset of items that most sharply distinguish between models, concentrating scarce annotation effort where it yields the greatest evaluative signal, rather than distributing it thinly across thousands of items of unequal informativeness. The recent \emph{tinyBenchmarks}~\cite{polo2024tinybenchmarks} is an example of this approach in practice. Complementary \textbf{active learning}~\cite{monarch2021human} strategies extend this logic iteratively: by selecting the next examples to annotate based on model uncertainty or expected information gain, annotation budgets can be directed toward the cases that most change our understanding of model behaviour. Together, these approaches offer a principled path to doing more with less, a direct response to the annotation scarcity constraint. The key limitation is that IRT calibration requires a  sufficient initial annotation pool, which remains a bootstrapping challenge for  the most severely under-resourced languages.
\end{itemize}

Each of these approaches involves significant trade-offs between scalability, validity, and equity. The current era is defined by the search for a balance between the technical hunger for massive datasets and the ethical  imperative of epistemic justice.

\section{Discussion: Implications for the Future of Low-Resource NLP}

Our survey reveals a field at a critical inflection point. The Annotation Scarcity Paradox is not merely a logistical inconvenience to be solved through cleverer algorithms; it is a structural feature of the evaluation ecosystem that shapes what can be known, and what counts as progress, in low-resource NLP\@. When the capacity to evaluate models trails so far behind the capacity to generate them, the field risks optimising for statistical illusions rather than genuine linguistic utility. 

\subsection{The Need for Sustainable and Participatory Evaluation}

A sustainable evaluation ecosystem must shift from transactional, extractive  data-gathering to relational, community-embedded capacity building~\cite{datadotorg2026digitisation}.
We must distribute the labour of annotation and assessment more equitably, invest heavily in the development of local evaluation expertise, and create governance structures that give language communities meaningful agency over how their languages are represented and assessed~\cite{jo2020lessons}. Evaluation must be reframed not as the final hurdle of model deployment, but as an ongoing dialogue with the language community~\cite{sloane2022participation}. This requires institutional commitment from funding bodies, universities, and research consortia to finance long-term digital infrastructure,  rather than merely funding short-term methodological innovation or raw compute.

\subsection{Recognising the Limits of Human Capacity and Enforcing Transparency}

The field can no longer afford to treat human annotation as an infinitely  renewable, frictionless resource. We must develop and enforce clearer norms for reporting the human resources involved in benchmark construction and  evaluation. Building on frameworks like Data Statements \cite{bender-friedman-2018-data}, researchers must explicitly document team size, annotator demographics,  compensation arrangements, and power dynamics. Furthermore, treating inter-annotator  disagreement simply as ``noise'' to be averaged out erases vital sociolinguistic  variation. Releasing annotator metrics and acknowledging inherent subjectivity \cite{prabhakaran-etal-2021-releasing} would enable better calibration of confidence  in reported results, supporting more realistic, grounded assessments of what  evaluation findings actually mean in a real-world context.

\subsection{Avenues for Future Research}

To navigate this paradox, the NLP community must re-evaluate its incentive 
structures. We identify several promising directions for future research:

\begin{itemize}
    \item \textbf{Pluralistic Evaluation Frameworks:} Developing methodologies that explicitly model annotator uncertainty, cultural subjectivity, and dialectal variation, rather than forcing a single, universal ``ground truth'' onto complex linguistic tasks.
    \item \textbf{Re-aligning Shared Tasks:} Designing community competitions that explicitly reward data provenance, annotation transparency, and ethical governance, rather than solely ranking submissions by automated benchmark performance.
    \item \textbf{Community-Owned Infrastructure:} Fostering long-term, community-embedded partnerships that move beyond the extractive dynamics of one-off annotation campaigns. NLP (and AI research in general) faces the same challenge as helicopter research, well documented in health research~\cite{abimbola2020will,crane2011scrambling}
and we need to move beyond WEIRD (Western Educated Industrialised Rich Democratic) context~\cite{henrich2010weirdest}.
Research should increasingly focus on building local data trusts~\cite{delacroix2019bottom,olorunju2024african,okorie2025s,okorie2025addressing,rajab2025esethu} and governed-access repositories that ensure data sovereignty outlives any single grant cycle or publication. 
    
Sustaining such infrastructure beyond individual grant cycles will require alternative economic models: endowment structures analogous to open-source software foundations, revenue-sharing agreements with commercial users of community-owned data, or explicit financing from multilateral development institutions. See the example of Karya in India~\cite{Perrigo_2023,okolo2024Moving}.
\end{itemize}

Ultimately, overcoming the Annotation Scarcity Paradox will not be achieved by asking how we can evaluate models faster, but by critically examining who has the power, resources, and mandate to evaluate them at all.

\subsection{Limitations and Scope}

We acknowledge that the African NLP context is more deeply represented in this survey than other low-resource language communities, including Southeast Asian and Indigenous Americas NLP ecosystems. This reflects the author's positionality and professional network, and is itself a manifestation of the inequitable concentration of expertise this paper critiques. 

Accordingly, the claims of the paradox are most directly evidenced in the African NLP context; we treat this as a well-documented instance of a broader structural pattern that we expect manifests in analogous, if locally distinct, forms in Southeast Asian and Indigenous Americas language ecosystems.

Work such as those in communities like AmericasNLP~\cite{mager-etal-2021-findings} and SEACrowd~\cite{lovenia-etal-2024-seacrowd} show analogous annotation scarcity dynamics arising from distinct institutional actors (NGOs, universities, state bodies) and data sovereignty frameworks. In Southeast Asia, SEA-HELM further identifies cultural diagnostics as a distinct evaluation pillar largely absent from global benchmarks~\cite{susanto2025sea}; in South Asia, IndicGenBench documents that even 29 relatively higher-resourced Indic languages exhibit systematic generation quality gaps under multilingual LLMs~\cite{l2024indicgenbench}; and AmericasNLI demonstrates near-chance zero-shot NLI performance for most indigenous Americas languages~\cite{ebrahimi2022americasnli}.
We invite researchers embedded in other regional ecosystems to extend and contest the framework offered here.

As a critical narrative review, this survey makes no claim to exhaustive
coverage; the works cited are representative rather than comprehensive.

\section{Navigating the Tension: A Pragmatic Call to Action}

Drawing on the author's experience as a practitioner and community member in this space, this section offers a pragmatic complement to the structural critique above. The observations below are offered as practitioner reflections inviting empirical contestation, not as reviewed claims.

Critiquing the extractive nature of modern scaling often inadvertently triggers a secondary risk: research paralysis. Confronted with the immense difficulty of the Annotation Scarcity Paradox, the complexities of epistemic governance, and the historical debt of data colonialism, researchers may feel that working on low-resource languages is too ethically fraught or logistically demanding to pursue. There is a palpable anxiety among practitioners who find themselves caught in a precarious middle ground, tasked with the technical imperative of getting their languages represented in global, state-of-the-art models, while bearing the profound, often exhausting responsibility of representing their communities with absolute sincerity and cultural integrity.

It is crucial to emphasise that the messy, human side of computing is not an impediment to natural language processing; it is the fundamental core of the discipline. The friction of community engagement should not drive researchers away from these problems, but rather redefine what a `successful' project looks like. For researchers entering this space, pragmatic engagement must take precedence over the pursuit of perfect, frictionless scale. 

Practically, this means embracing `slow AI'~\cite{sambasivan2021everyone}, choosing to build a single, deeply verified, community-owned dataset that serves a specific local need, rather than feeling pressured to scrape millions of unverified tokens just to appear on a global leaderboard. It means accepting that building trust is non-linear, and that navigating the human dynamics of data collection is just as scientifically rigorous and valuable as optimising a model's loss function. 

Finally, a fundamental humility is required: we must acknowledge that the researchers advocating for and adopting these participatory frameworks are themselves imperfect. Missteps in cultural translation, logistical oversights in compensation, and the inadvertent replication of historic power dynamics will still inevitably occur. Yet, committing to this difficult path, despite its friction and our own fallibility, is absolutely essential for the long-term sustainability of our field. Embracing the human complexity of NLP is ultimately the only viable mechanism to earn, and rightfully keep, the trust of the communities we claim to serve.

\section{Conclusion}
 
The last decade of low-resource NLP has produced genuine and important progress. Yet the pace of benchmark creation, and the ever-increasing data hunger of modern generative scaling, has vastly outrun the human  infrastructure needed to support it. This imbalance has produced a profound gap between reported machine performance and our warranted confidence in that  performance. We have characterised this gap as the Annotation Scarcity Paradox and traced its emergence across three distinct phases of the  field's development, arguing that this bottleneck is not merely logistical, but deeply structural.

We call on the wider NLP community to critically examine the assumptions underlying current scaling practices and to transition away from transactional  data extraction. We must invest, collectively, in more sustainable, equitable,  and epistemically honest approaches that prioritise shared ownership and  governed access. 

Embracing this complexity, and accepting the inherent friction, slow pace,  and imperfection of the human work required to achieve it, is no longer optional. Doing so is not only a fundamental requirement for scientific  rigour; it is a matter of epistemic justice toward the language communities whose linguistic heritage and communicative needs this research ultimately  serves.
 
\appendix
 
\section*{Ethical Statement}
 
This paper presents a survey and does not involve human subjects research,
the collection of new data, or the deployment of systems. The survey
addresses questions of equity and community involvement in NLP research;
we have endeavoured to engage with these questions with appropriate care
and humility.
 
\section*{Acknowledgments}

This work is supported by the ABSA Chair of Data Science at the University
of Pretoria, and has benefited from gifts from NVIDIA, Google.org,
OpenAI, and Meta, as well as funding from the UK International Development
programme and IDRC Ottawa (AI4D Africa). This paper was produced through
an Oppenheimer Memorial Trust (OMT) Sabbatical Study Grant awarded to the
author. The author thanks Dr.\ Alvitta Ottley for her input in refining
Figure~\ref{fig:africa-funnel}.

\bibliographystyle{named}
\bibliography{ijcai26}

\begin{thebibliography}{}

\bibitem[\protect\citeauthoryear{Abbott and Martinus}{2018}]{abbott2018towards}
Jade~Z Abbott and Laura Martinus.
\newblock Towards neural machine translation for african languages.
\newblock {\em arXiv preprint arXiv:1811.05467}, 2018.

\bibitem[\protect\citeauthoryear{Abdulmumin \bgroup \em et al.\egroup
  }{2024}]{abdulmumin-etal-2024-correcting}
Idris Abdulmumin, Sthembiso Mkhwanazi, Mahlatse Mbooi, Shamsuddeen~Hassan
  Muhammad, Ibrahim~Said Ahmad, Neo Putini, Miehleketo Mathebula, Matimba
  Shingange, Tajuddeen Gwadabe, and Vukosi Marivate.
\newblock Correcting {FLORES} evaluation dataset for four {A}frican languages.
\newblock In Barry Haddow, Tom Kocmi, Philipp Koehn, and Christof Monz,
  editors, {\em Proceedings of the Ninth Conference on Machine Translation},
  pages 570--578, Miami, Florida, USA, November 2024. Association for
  Computational Linguistics.

\bibitem[\protect\citeauthoryear{Abimbola and Pai}{2020}]{abimbola2020will}
Seye Abimbola and Madhukar Pai.
\newblock Will global health survive its decolonisation?
\newblock {\em The Lancet}, 396(10263):1627--1628, 2020.

\bibitem[\protect\citeauthoryear{Adams \bgroup \em et al.\egroup
  }{2017}]{adams2017cross}
Oliver Adams, Adam Makarucha, Graham Neubig, Steven Bird, and Trevor Cohn.
\newblock Cross-lingual word embeddings for low-resource language modeling.
\newblock In {\em Proceedings of the 15th Conference of the European Chapter of
  the Association for Computational Linguistics: Volume 1, Long Papers}, pages
  937--947, 2017.

\bibitem[\protect\citeauthoryear{Adebara}{2025}]{adebara2025ai}
Ife Adebara.
\newblock {AI} and language data flaring in {A}frica: Addressing the
  low-resource challenge.
\newblock {\em Policy Brief No. 216}, 2025.

\bibitem[\protect\citeauthoryear{Adelani \bgroup \em et al.\egroup
  }{2022}]{adelani2022masakhaner}
David~Ifeoluwa Adelani, Graham Neubig, Sebastian Ruder, Shruti Rijhwani,
  Michael Beukman, Chester Palen-Michel, Constantine Lignos, Jesujoba Alabi,
  Shamsuddeen~H Muhammad, Peter Nabende, et~al.
\newblock Masakhaner 2.0: Africa-centric transfer learning for named entity
  recognition.
\newblock In {\em Proceedings of the 2022 Conference on Empirical Methods in
  Natural Language Processing}, pages 4488--4508, 2022.

\bibitem[\protect\citeauthoryear{Adelani \bgroup \em et al.\egroup
  }{2025}]{adelani2025irokobench}
David~Ifeoluwa Adelani, Jessica Ojo, Israel~Abebe Azime, Jian~Yun Zhuang,
  Jesujoba Alabi, Xuanli He, Millicent Ochieng, Sara Hooker, Andiswa Bukula,
  En-Shiun~Annie Lee, et~al.
\newblock Irokobench: A new benchmark for african languages in the age of large
  language models.
\newblock In {\em Proceedings of the 2025 Conference of the Nations of the
  Americas Chapter of the Association for Computational Linguistics: Human
  Language Technologies (Volume 1: Long Papers)}, pages 2732--2757, 2025.

\bibitem[\protect\citeauthoryear{Agi{\'c} and Vuli{\'c}}{2019}]{agic2019jw300}
{\v{Z}}eljko Agi{\'c} and Ivan Vuli{\'c}.
\newblock {JW}300: A wide-coverage parallel corpus for low-resource languages.
\newblock In {\em Proceedings of the 57th Annual Meeting of the Association for
  Computational Linguistics}, pages 3204--3210, 2019.

\bibitem[\protect\citeauthoryear{Ahuja \bgroup \em et al.\egroup
  }{2023}]{ahuja2023mega}
Kabir Ahuja, Harshita Diddee, Rishav Hada, Millicent Ochieng, Krithika Ramesh,
  Prachi Jain, Akshay Nambi, Tanuja Ganu, Sameer Segal, Mohamed Ahmed, et~al.
\newblock Mega: Multilingual evaluation of generative ai.
\newblock In {\em Proceedings of the 2023 Conference on Empirical Methods in
  Natural Language Processing}, pages 4232--4267, 2023.

\bibitem[\protect\citeauthoryear{Alabi \bgroup \em et al.\egroup
  }{2022}]{alabi2022adapting}
Jesujoba Alabi, David~Ifeoluwa Adelani, Marius Mosbach, and Dietrich Klakow.
\newblock Adapting pre-trained language models to {A}frican languages via
  multilingual adaptive fine-tuning.
\newblock In {\em Proceedings of the 29th International Conference on
  Computational Linguistics}, pages 4336--4349, 2022.

\bibitem[\protect\citeauthoryear{Alabi \bgroup \em et al.\egroup
  }{2025}]{alabi2025charting}
Jesujoba Alabi, Michael~A Hedderich, David~Ifeoluwa Adelani, and Dietrich
  Klakow.
\newblock Charting the landscape of african nlp: Mapping progress and shaping
  the road ahead.
\newblock In {\em Proceedings of the 2025 Conference on Empirical Methods in
  Natural Language Processing}, pages 27795--27829, 2025.

\bibitem[\protect\citeauthoryear{Ardila \bgroup \em et al.\egroup
  }{2020}]{ardila2020common}
Rosana Ardila, Megan Branson, Kelly Davis, Michael Kohler, Josh Meyer, Michael
  Henretty, Reuben Morais, Lindsay Saunders, Francis Tyers, and Gregor Weber.
\newblock Common voice: A massively-multilingual speech corpus.
\newblock In {\em Proceedings of the twelfth language resources and evaluation
  conference}, pages 4218--4222, 2020.

\bibitem[\protect\citeauthoryear{Belay \bgroup \em et al.\egroup
  }{2025}]{belay2025rise}
Tadesse~Destaw Belay, Kedir~Yassin Hussen, Sukairaj~Hafiz Imam, Ibrahim~Said
  Ahmad, Isa Inuwa-Dutse, Abrham~Belete Haile, Grigori Sidorov, Iqra Ameer,
  Idris Abdulmumin, Tajuddeen Gwadabe, et~al.
\newblock The rise of africanlp: Contributions, contributors, and community
  impact (2005-2025).
\newblock {\em arXiv preprint arXiv:2509.25477}, 2025.

\bibitem[\protect\citeauthoryear{Bender and
  Friedman}{2018}]{bender-friedman-2018-data}
Emily~M. Bender and Batya Friedman.
\newblock Data statements for natural language processing: Toward mitigating
  system bias and enabling better science.
\newblock {\em Transactions of the Association for Computational Linguistics},
  6:587--604, 2018.

\bibitem[\protect\citeauthoryear{Bender \bgroup \em et al.\egroup
  }{2021}]{bender2021dangers}
Emily~M Bender, Timnit Gebru, Angelina McMillan-Major, and Shmargaret
  Shmitchell.
\newblock On the dangers of stochastic parrots: Can language models be too big?
\newblock In {\em Proceedings of the 2021 ACM conference on fairness,
  accountability, and transparency}, pages 610--623, 2021.

\bibitem[\protect\citeauthoryear{Benjamin}{2019}]{benjamin2019race}
Ruha Benjamin.
\newblock {\em Race After Technology: Abolitionist Tools for the New Jim Code}.
\newblock Polity, Medford, MA, 2019.

\bibitem[\protect\citeauthoryear{Bird}{2020}]{bird2020decolonising}
Steven Bird.
\newblock Decolonising speech and language technology.
\newblock In {\em Proceedings of the 28th international conference on
  computational linguistics}, pages 3504--3519, 2020.

\bibitem[\protect\citeauthoryear{Birhane and Prabhu}{2021}]{birhane2021large}
Abeba Birhane and Vinay~Uday Prabhu.
\newblock Large image datasets: A pyrrhic win for computer vision?
\newblock In {\em 2021 IEEE Winter Conference on Applications of Computer
  Vision (WACV)}, pages 1536--1546. IEEE, 2021.

\bibitem[\protect\citeauthoryear{Birhane \bgroup \em et al.\egroup
  }{2022}]{birhane2022values}
Abeba Birhane, Pratyusha Kalluri, Dallas Card, William Agnew, Ravit Dotan, and
  Michelle Bao.
\newblock The values encoded in machine learning research.
\newblock In {\em Proceedings of the 2022 ACM conference on fairness,
  accountability, and transparency}, pages 173--184, 2022.

\bibitem[\protect\citeauthoryear{Birhane}{2020}]{birhane2020algorithmic}
Abeba Birhane.
\newblock Algorithmic colonization of africa.
\newblock {\em SCRIPTed}, 17:389, 2020.

\bibitem[\protect\citeauthoryear{Blasi \bgroup \em et al.\egroup
  }{2022}]{blasi2022systematic}
Damian Blasi, Antonios Anastasopoulos, and Graham Neubig.
\newblock Systematic inequalities in language technology performance across the
  world’s languages.
\newblock In {\em Proceedings of the 60th Annual Meeting of the Association for
  Computational Linguistics (Volume 1: Long Papers)}, pages 5486--5505, 2022.

\bibitem[\protect\citeauthoryear{Carroll \bgroup \em et al.\egroup
  }{2023}]{carroll2023care}
Stephanie~Russo Carroll, Ibrahim Garba, Oscar~L Figueroa-Rodr{\'\i}guez, Jarita
  Holbrook, Raymond Lovett, Simeon Materechera, Mark Parsons, Kay Raseroka,
  Desi Rodriguez-Lonebear, Robyn Rowe, et~al.
\newblock The care principles for indigenous data governance.
\newblock {\em Open Scholarship Press Curated Volumes: Policy}, 2023.

\bibitem[\protect\citeauthoryear{Chen \bgroup \em et al.\egroup
  }{2023}]{chen2023empirical}
Jiaao Chen, Derek Tam, Colin Raffel, Mohit Bansal, and Diyi Yang.
\newblock An empirical survey of data augmentation for limited data learning in
  nlp.
\newblock {\em Transactions of the Association for Computational Linguistics},
  11:191--211, 2023.

\bibitem[\protect\citeauthoryear{Chiu \bgroup \em et al.\egroup
  }{2025}]{chiu2025culturalbench}
Yu~Ying Chiu, Liwei Jiang, Bill~Yuchen Lin, Chan~Young Park, Shuyue~Stella Li,
  Sahithya Ravi, Mehar Bhatia, Maria Antoniak, Yulia Tsvetkov, Vered Shwartz,
  et~al.
\newblock Culturalbench: A robust, diverse and challenging benchmark for
  measuring lms’ cultural knowledge through human-ai red-teaming.
\newblock In {\em Proceedings of the 63rd Annual Meeting of the Association for
  Computational Linguistics (Volume 1: Long Papers)}, pages 25663--25701, 2025.

\bibitem[\protect\citeauthoryear{Conneau \bgroup \em et al.\egroup
  }{2020}]{conneau2020unsupervised}
Alexis Conneau, Kartikay Khandelwal, Naman Goyal, Vishrav Chaudhary, Guillaume
  Wenzek, Francisco Guzm{\'a}n, Edouard Grave, Myle Ott, Luke Zettlemoyer, and
  Veselin Stoyanov.
\newblock Unsupervised cross-lingual representation learning at scale.
\newblock In {\em Proceedings of the 58th annual meeting of the association for
  computational linguistics}, pages 8440--8451, 2020.

\bibitem[\protect\citeauthoryear{Couldry and Mejias}{2019}]{couldry2019costs}
Nick Couldry and Ulises~A. Mejias.
\newblock {\em The Costs of Connection: How Data Is Colonizing Human Life and
  Appropriating It for Capitalism}.
\newblock Stanford University Press, Stanford, CA, 2019.

\bibitem[\protect\citeauthoryear{Crane}{2011}]{crane2011scrambling}
Johanna Crane.
\newblock Scrambling for africa? universities and global health.
\newblock {\em The Lancet}, 377(9775):1388--1390, 2011.

\bibitem[\protect\citeauthoryear{{DataDotOrg}}{2026}]{datadotorg2026digitisation}
{DataDotOrg}.
\newblock Digitisation of oral data for nlp of low-resource languages:
  Practical methods and processes for scalable and sustainable ecosystem
  development.
\newblock Playbook, DataDotOrg, Washington, D.C., USA, 2026.
\newblock A playbook for building sustainable African language technology
  ecosystems.

\bibitem[\protect\citeauthoryear{de Wet \bgroup \em et al.\egroup
  }{2022}]{de2022localising}
Febe de~Wet, Andiswa Bukula, Willem Karsten, Martin Puttkammer, Erwin
  Schillack, Rone Wierenga, and Roald Eiselen.
\newblock Localising the mozilla common voice platform for south africa’s
  official languages.
\newblock {\em Journal of the Digital Humanities Association of Southern Africa
  (DHASA)}, 4(01), 2022.

\bibitem[\protect\citeauthoryear{Delacroix and
  Lawrence}{2019}]{delacroix2019bottom}
Sylvie Delacroix and Neil~D Lawrence.
\newblock Bottom-up data trusts: Disturbing the ‘one size fits all’approach
  to data governance.
\newblock {\em International data privacy law}, 9(4):236--252, 2019.

\bibitem[\protect\citeauthoryear{Dencik \bgroup \em et al.\egroup
  }{2019}]{dencik2019exploring}
Lina Dencik, Arne Hintz, Joanna Redden, and Emiliano Trer{\'e}.
\newblock Exploring data justice: Conceptions, applications and directions.
\newblock {\em Information, Communication \& Society}, 22(7):873--881, 2019.

\bibitem[\protect\citeauthoryear{Devlin \bgroup \em et al.\egroup
  }{2019}]{devlin2019bert}
Jacob Devlin, Ming-Wei Chang, Kenton Lee, and Kristina Toutanova.
\newblock Bert: Pre-training of deep bidirectional transformers for language
  understanding.
\newblock In {\em Proceedings of the 2019 conference of the North American
  chapter of the association for computational linguistics: human language
  technologies, volume 1 (long and short papers)}, pages 4171--4186, 2019.

\bibitem[\protect\citeauthoryear{Dhole \bgroup \em et al.\egroup
  }{2023}]{dhole2023nl}
Kaustubh Dhole, Varun Gangal, Sebastian Gehrmann, Aadesh Gupta, Zhenhao Li,
  Saad Mahamood, Abinaya Mahadiran, Simon Mille, Ashish Shrivastava, Samson
  Tan, et~al.
\newblock Nl-augmenter: A framework for task-sensitive natural language
  augmentation.
\newblock {\em Northern European Journal of Language Technology}, 9, 2023.

\bibitem[\protect\citeauthoryear{Eberhard \bgroup \em et al.\egroup
  }{2025}]{eberhard2025ethnologue}
David~M. Eberhard, Gary~F. Simons, and Charles~D. Fennig.
\newblock {Ethnologue}: Languages of the world.
\newblock SIL International, 2025.

\bibitem[\protect\citeauthoryear{Ebrahimi \bgroup \em et al.\egroup
  }{2022}]{ebrahimi2022americasnli}
Abteen Ebrahimi, Manuel Mager, Adam Wiemerslage, Pavel Denisov, Katharina Kann,
  et~al.
\newblock {AmericasNLI}: Evaluating zero-shot natural language understanding of
  pretrained multilingual models in truly low-resource languages.
\newblock In {\em Proceedings of the 60th Annual Meeting of the Association for
  Computational Linguistics (Volume 1: Long Papers)}, pages 6279--6299, 2022.

\bibitem[\protect\citeauthoryear{Effoduh}{2026}]{effoduh2026decolonizing}
Jake~Okechukwu Effoduh.
\newblock Decolonizing the governance of artificial intelligence in africa:
  from normative mimicry to epistemic sovereignty.
\newblock {\em Science and Public Policy}, 53(2):245--257, 2026.

\bibitem[\protect\citeauthoryear{Eiselen and
  Puttkammer}{2014}]{eiselen2014developing}
Roald Eiselen and Martin~J Puttkammer.
\newblock Developing text resources for ten {S}outh {A}frican languages.
\newblock In {\em Proceedings of the Ninth International Conference on Language
  Resources and Evaluation (LREC'14)}, pages 3698--3703, 2014.

\bibitem[\protect\citeauthoryear{Feng \bgroup \em et al.\egroup
  }{2021}]{feng2021survey}
Steven~Y Feng, Varun Gangal, Jason Wei, Sarath Chandar, Soroush Vosoughi,
  Teruko Mitamura, and Eduard Hovy.
\newblock A survey of data augmentation approaches for nlp.
\newblock In {\em Findings of the association for computational linguistics:
  ACL-IJCNLP 2021}, pages 968--988, 2021.

\bibitem[\protect\citeauthoryear{Grant and Booth}{2009}]{grant2009typology}
Maria~J Grant and Andrew Booth.
\newblock A typology of reviews: an analysis of 14 review types and associated
  methodologies.
\newblock {\em Health information \& libraries journal}, 26(2):91--108, 2009.

\bibitem[\protect\citeauthoryear{Gu \bgroup \em et al.\egroup
  }{2018}]{gu2018universal}
Jiatao Gu, Hany~Hassan Awadalla, Jacob Devlin, and Victor~OK Li.
\newblock Universal neural machine translation for extremely low resource
  languages.
\newblock In {\em Proceedings of the 2018 Conference of the North American
  Chapter of the Association for Computational Linguistics: Human Language
  Technologies, Volume 1 (Long Papers)}, pages 344--354, 2018.

\bibitem[\protect\citeauthoryear{Henrich \bgroup \em et al.\egroup
  }{2010}]{henrich2010weirdest}
Joseph Henrich, Steven~J Heine, and Ara Norenzayan.
\newblock The weirdest people in the world?
\newblock {\em Behavioral and Brain Sciences}, 33(2-3):61--83, 2010.

\bibitem[\protect\citeauthoryear{Hershcovich \bgroup \em et al.\egroup
  }{2022}]{hershcovich2022challenges}
Daniel Hershcovich, Stella Frank, Heather Lent, Miryam De~Lhoneux, Mostafa
  Abdou, Stephanie Brandl, Emanuele Bugliarello, Laura~Cabello Piqueras, Ilias
  Chalkidis, Ruixiang Cui, et~al.
\newblock Challenges and strategies in cross-cultural nlp.
\newblock In {\em Proceedings of the 60th Annual Meeting of the Association for
  Computational Linguistics (Volume 1: Long Papers)}, pages 6997--7013, 2022.

\bibitem[\protect\citeauthoryear{Hu \bgroup \em et al.\egroup
  }{2020}]{hu2020xtreme}
Junjie Hu, Sebastian Ruder, Aditya Siddhant, Graham Neubig, Orhan Firat, and
  Melvin Johnson.
\newblock {XTREME}: A massively multilingual multi-task benchmark for
  evaluating cross-lingual generalisation.
\newblock In {\em International conference on machine learning}, pages
  4411--4421. PMLR, 2020.

\bibitem[\protect\citeauthoryear{Iliadis and Russo}{2016}]{iliadis2016critical}
Andrew Iliadis and Federica Russo.
\newblock Critical data studies: An introduction.
\newblock {\em Big Data \& Society}, 3(2), 2016.

\bibitem[\protect\citeauthoryear{Jo and Gebru}{2020}]{jo2020lessons}
Eun~Seo Jo and Timnit Gebru.
\newblock Lessons from archives: Strategies for collecting sociocultural data
  in machine learning.
\newblock In {\em Proceedings of the 2020 conference on fairness,
  accountability, and transparency}, pages 306--316, 2020.

\bibitem[\protect\citeauthoryear{Joshi \bgroup \em et al.\egroup
  }{2020}]{joshi2020state}
Pratik Joshi, Sebastin Santy, Amar Budhiraja, Kalika Bali, and Monojit
  Choudhury.
\newblock The state and fate of linguistic diversity and inclusion in the nlp
  world.
\newblock In {\em Proceedings of the 58th annual meeting of the association for
  computational linguistics}, pages 6282--6293, 2020.

\bibitem[\protect\citeauthoryear{Kholodna \bgroup \em et al.\egroup
  }{2024}]{kholodna2024llms}
Nataliia Kholodna, Sahib Julka, Mohammad Khodadadi, Muhammed~Nurullah Gumus,
  and Michael Granitzer.
\newblock Llms in the loop: Leveraging large language model annotations for
  active learning in low-resource languages.
\newblock In {\em Joint European Conference on Machine Learning and Knowledge
  Discovery in Databases}, pages 397--412. Springer, 2024.

\bibitem[\protect\citeauthoryear{King}{2015}]{king2015practical}
Benjamin~Philip King.
\newblock {\em Practical Natural Language Processing for Low-Resource
  Languages}.
\newblock PhD thesis, University of Michigan, 2015.

\bibitem[\protect\citeauthoryear{Klie \bgroup \em et al.\egroup
  }{2023}]{klie-etal-2023-lessons}
Jan-Christoph Klie, Ji-Ung Lee, Kevin Stowe, G{\"o}zde {\c{S}}ahin,
  Nafise~Sadat Moosavi, Luke Bates, Dominic Petrak, Richard Eckart De~Castilho,
  and Iryna Gurevych.
\newblock Lessons learned from a citizen science project for natural language
  processing.
\newblock In Andreas Vlachos and Isabelle Augenstein, editors, {\em Proceedings
  of the 17th Conference of the European Chapter of the Association for
  Computational Linguistics}, pages 3594--3608, Dubrovnik, Croatia, May 2023.
  Association for Computational Linguistics.

\bibitem[\protect\citeauthoryear{Kunchukuttan \bgroup \em et al.\egroup
  }{2018}]{kunchukuttan2018iit}
Anoop Kunchukuttan, Pratik Mehta, and Pushpak Bhattacharyya.
\newblock The {IIT} {B}ombay {E}nglish-{H}indi parallel corpus.
\newblock In {\em Proceedings of the Eleventh International Conference on
  Language Resources and Evaluation (LREC 2018)}, 2018.

\bibitem[\protect\citeauthoryear{Lalor \bgroup \em et al.\egroup
  }{2019}]{lalor-etal-2019-learning}
John~P. Lalor, Hao Wu, and Hong Yu.
\newblock Learning latent parameters without human response patterns: Item
  response theory with artificial crowds.
\newblock In {\em Proceedings of the 2019 Conference on Empirical Methods in
  Natural Language Processing (EMNLP-IJCNLP)}, pages 4674--4684, Hong Kong,
  China, November 2019. Association for Computational Linguistics.

\bibitem[\protect\citeauthoryear{Lovenia \bgroup \em et al.\egroup
  }{2024}]{lovenia-etal-2024-seacrowd}
Holy Lovenia, Rahmad Mahendra, Salsabil~Maulana Akbar, Lester James~V. Miranda,
  Jennifer Santoso, Elyanah Aco, Akhdan Fadhilah, Jonibek Mansurov,
  Joseph~Marvin Imperial, Onno~P. Kampman, Joel Ruben~Antony Moniz, Muhammad
  Ravi~Shulthan Habibi, Frederikus Hudi, Railey Montalan, Ryan Ignatius,
  Joanito~Agili Lopo, William Nixon, B{\"o}rje~F. Karlsson, James Jaya,
  Ryandito Diandaru, Yuze Gao, Patrick Amadeus, Bin Wang, Jan Christian~Blaise
  Cruz, Chenxi Whitehouse, Ivan~Halim Parmonangan, Maria Khelli, Wenyu Zhang,
  Lucky Susanto, Reynard~Adha Ryanda, Sonny~Lazuardi Hermawan, Dan~John
  Velasco, Muhammad Dehan~Al Kautsar, Willy~Fitra Hendria, Yasmin Moslem, Noah
  Flynn, Muhammad~Farid Adilazuarda, Haochen Li, Johanes Lee, R.~Damanhuri,
  Shuo Sun, Muhammad~Reza Qorib, Amirbek Djanibekov, Wei~Qi Leong, Quyet~V. Do,
  Niklas Muennighoff, Tanrada Pansuwan, Ilham~Firdausi Putra, Yan Xu, Tai~Ngee
  Chia, Ayu Purwarianti, Sebastian Ruder, William Tjhi, Peerat Limkonchotiwat,
  Alham~Fikri Aji, Sedrick Keh, Genta~Indra Winata, Ruochen Zhang, Fajri Koto,
  Zheng-Xin Yong, and Samuel Cahyawijaya.
\newblock {SEAC}rowd: A multilingual multimodal data hub and benchmark suite
  for {S}outheast {A}sian languages.
\newblock In Yaser Al-Onaizan, Mohit Bansal, and Yun-Nung Chen, editors, {\em
  Proceedings of the 2024 Conference on Empirical Methods in Natural Language
  Processing}, pages 5155--5203, Miami, Florida, USA, November 2024.
  Association for Computational Linguistics.

\bibitem[\protect\citeauthoryear{Mager \bgroup \em et al.\egroup
  }{2018}]{mager2018challenges}
Manuel Mager, Ximena Gutierrez-Vasques, Gerardo Sierra, and Ivan Meza-Ruiz.
\newblock Challenges of language technologies for the indigenous languages of
  the {A}mericas.
\newblock In {\em Proceedings of the 27th International Conference on
  Computational Linguistics}, pages 55--69, 2018.

\bibitem[\protect\citeauthoryear{Mager \bgroup \em et al.\egroup
  }{2021}]{mager-etal-2021-findings}
Manuel Mager, Arturo Oncevay, Abteen Ebrahimi, John Ortega, Annette Rios,
  Angela Fan, Ximena Gutierrez-Vasques, Luis Chiruzzo, Gustavo
  Gim{\'e}nez-Lugo, Ricardo Ramos, Ivan~Vladimir Meza~Ruiz, Rolando
  Coto-Solano, Alexis Palmer, Elisabeth Mager-Hois, Vishrav Chaudhary, Graham
  Neubig, Ngoc~Thang Vu, and Katharina Kann.
\newblock Findings of the {A}mericas{NLP} 2021 shared task on open machine
  translation for indigenous languages of the {A}mericas.
\newblock In Manuel Mager, Arturo Oncevay, Annette Rios, Ivan Vladimir~Meza
  Ruiz, Alexis Palmer, Graham Neubig, and Katharina Kann, editors, {\em
  Proceedings of the First Workshop on Natural Language Processing for
  Indigenous Languages of the Americas}, pages 202--217, Online, June 2021.
  Association for Computational Linguistics.

\bibitem[\protect\citeauthoryear{Manduchi \bgroup \em et al.\egroup
  }{2024}]{manduchi2024challenges}
Laura Manduchi, Clara Meister, Kushagra Pandey, Robert Bamler, Ryan Cotterell,
  Sina D{\"a}ubener, Sophie Fellenz, Asja Fischer, Thomas G{\"a}rtner, Matthias
  Kirchler, et~al.
\newblock On the challenges and opportunities in generative ai.
\newblock {\em arXiv preprint arXiv:2403.00025}, 2024.

\bibitem[\protect\citeauthoryear{Microsoft
  Research~Africa}{2026}]{pazabench2026}
Nairobi Microsoft Research~Africa.
\newblock {PazaBench}: A benchmark for automatic speech recognition on low
  resource languages.
\newblock
  \url{https://www.microsoft.com/en-us/research/project/project-gecko/}, 2026.
\newblock Alpha version. Part of Project Gecko - Equitable Generative AI for
  the Global Majority.

\bibitem[\protect\citeauthoryear{Monarch}{2021}]{monarch2021human}
Robert~Munro Monarch.
\newblock {\em Human-in-the-Loop Machine Learning: Active learning and
  annotation for human-centered AI}.
\newblock Simon and Schuster, 2021.

\bibitem[\protect\citeauthoryear{Muhammad \bgroup \em et al.\egroup
  }{2023}]{muhammad-etal-2023-afrisenti}
Shamsuddeen~Hassan Muhammad, Idris Abdulmumin, Abinew~Ali Ayele, Nedjma
  Ousidhoum, David~Ifeoluwa Adelani, Seid~Muhie Yimam, Ibrahim~Sa'id Ahmad,
  Meriem Beloucif, Saif~M. Mohammad, Sebastian Ruder, Oumaima Hourrane, Pavel
  Brazdil, Alipio Jorge, Felermino D{\'a}rio M{\'a}rio~Ant{\'o}nio Ali, Davis
  David, Salomey Osei, Bello Shehu~Bello, Falalu Ibrahim, Tajuddeen Gwadabe,
  Samuel Rutunda, Tadesse Belay, Wendimu~Baye Messelle, Hailu~Beshada Balcha,
  Sisay~Adugna Chala, Hagos~Tesfahun Gebremichael, Bernard Opoku, and Stephen
  Arthur.
\newblock {A}fri{S}enti: A {T}witter sentiment analysis benchmark for {A}frican
  languages.
\newblock In Houda Bouamor, Juan Pino, and Kalika Bali, editors, {\em
  Proceedings of the 2023 Conference on Empirical Methods in Natural Language
  Processing}, pages 13968--13981, Singapore, December 2023. Association for
  Computational Linguistics.

\bibitem[\protect\citeauthoryear{Nekoto \bgroup \em et al.\egroup
  }{2020}]{nekoto2020participatory}
Wilhelmina Nekoto, Vukosi Marivate, Tshinondiwa Matsila, Timi Fasubaa, Taiwo
  Fagbohungbe, Solomon~Oluwole Akinola, Shamsuddeen Muhammad, Salomon~Kabongo
  Kabenamualu, Salomey Osei, Freshia Sackey, et~al.
\newblock Participatory research for low-resourced machine translation: A case
  study in {A}frican languages.
\newblock In {\em Findings of the Association for Computational Linguistics:
  EMNLP 2020}, pages 2144--2160, 2020.

\bibitem[\protect\citeauthoryear{Ojo \bgroup \em et al.\egroup
  }{2025}]{ojo2025afrobench}
Jessica Ojo, Odunayo Ogundepo, Akintunde Oladipo, Kelechi Ogueji, Jimmy Lin,
  Pontus Stenetorp, and David~Ifeoluwa Adelani.
\newblock Afrobench: how good are large language models on african languages?
\newblock In {\em Findings of the Association for Computational Linguistics:
  ACL 2025}, pages 19048--19095, 2025.

\bibitem[\protect\citeauthoryear{Okolo and Tano}{2024}]{okolo2024Moving}
Chinasa Okolo and Marie Tano.
\newblock Moving toward truly responsible {AI} development in the global {AI}
  market, 2024.
\newblock Brookings Institution.

\bibitem[\protect\citeauthoryear{Okolo}{2024}]{okolo2024reforming}
Chinasa Okolo.
\newblock Reforming data regulation to advance {AI} governance in {Africa},
  2024.

\bibitem[\protect\citeauthoryear{Okorie and Omino}{2025}]{okorie2025addressing}
Chijioke Okorie and Melissa Omino.
\newblock Addressing inequitable openness in licences for sharing african data
  and datasets through the nwulite obodo open data licence.
\newblock {\em Law, Tech. \& Hum.}, 7:94, 2025.

\bibitem[\protect\citeauthoryear{Okorie}{2025}]{okorie2025s}
Chijioke Okorie.
\newblock It’s the noodl license--awesome and amazingly geeky!
\newblock {\em Available at SSRN 5339254}, 2025.

\bibitem[\protect\citeauthoryear{Olorunju and
  Adams}{2024}]{olorunju2024african}
Nokuthula Olorunju and Rachel Adams.
\newblock African data trusts: new tools towards collective data governance?
\newblock {\em Information \& Communications Technology Law}, 33(1):85--98,
  2024.

\bibitem[\protect\citeauthoryear{{Omnilingual ASR team} \bgroup \em et
  al.\egroup }{2025}]{meta2025omnilingual}
{Omnilingual ASR team}, Gil Keren, Artyom Kozhevnikov, Yen Meng, Christophe
  Ropers, Matthew Setzler, Skyler Wang, Ife Adebara, Michael Auli, Can
  Balioglu, Kevin Chan, Chierh Cheng, Joe Chuang, Caley Droof, Mark
  Duppenthaler, Paul-Ambroise Duquenne, Alexander Erben, Cynthia Gao,
  Gabriel~Mejia Gonzalez, Kehan Lyu, Sagar Miglani, Vineel Pratap, Kaushik~Ram
  Sadagopan, Safiyyah Saleem, Arina Turkatenko, Albert Ventayol-Boada,
  Zheng-Xin Yong, Yu-An Chung, Jean Maillard, Rashel Moritz, Alexandre
  Mourachko, Mary Williamson, and Shireen Yates.
\newblock Omnilingual asr: Open-source multilingual speech recognition for
  1600+ languages.
\newblock {\em arXiv preprint arXiv: 2511.09690}, 2025.

\bibitem[\protect\citeauthoryear{Orlic}{2021}]{orlic2021outreach}
Davor Orlic.
\newblock Outreach programme to strengthen the {AI4D} network: final technical
  report.
\newblock Technical report, AI4D Africa, 2021.

\bibitem[\protect\citeauthoryear{Perrigo}{2023}]{Perrigo_2023}
Billy Perrigo.
\newblock Ai by the people, for the people, July 2023.

\bibitem[\protect\citeauthoryear{Polo \bgroup \em et al.\egroup
  }{2024}]{polo2024tinybenchmarks}
Felipe~Maia Polo, Lucas Weber, Leshem Choshen, Yuekai Sun, Gongjun Xu, and
  Mikhail Yurochkin.
\newblock tinybenchmarks: evaluating llms with fewer examples.
\newblock In {\em Proceedings of the 41st International Conference on Machine
  Learning}, pages 34303--34326, 2024.

\bibitem[\protect\citeauthoryear{Prabhakaran \bgroup \em et al.\egroup
  }{2021}]{prabhakaran-etal-2021-releasing}
Vinodkumar Prabhakaran, Aida Mostafazadeh~Davani, and Mark Diaz.
\newblock On releasing annotator-level labels and information in datasets.
\newblock In Claire Bonial and Nianwen Xue, editors, {\em Proceedings of the
  Joint 15th Linguistic Annotation Workshop (LAW) and 3rd Designing Meaning
  Representations (DMR) Workshop}, pages 133--138, Punta Cana, Dominican
  Republic, November 2021. Association for Computational Linguistics.

\bibitem[\protect\citeauthoryear{Rajab \bgroup \em et al.\egroup
  }{2025}]{rajab2025esethu}
Jenalea Rajab, Anuoluwapo Aremu, Everlyn~Asiko Chimoto, Dale Dunbar, Graham
  Morrissey, Fadel Thior, Luandrie Potgieter, Jessica Ojo, Atnafu~Lambebo
  Tonja, Wilhelmina~NdapewaOnyothi Nekoto, et~al.
\newblock The esethu framework: Reimagining sustainable dataset governance and
  curation for low-resource languages.
\newblock In {\em Proceedings of the 63rd Annual Meeting of the Association for
  Computational Linguistics (Volume 1: Long Papers)}, pages 30763--30776, 2025.

\bibitem[\protect\citeauthoryear{Rodriguez \bgroup \em et al.\egroup
  }{2021}]{rodriguez-etal-2021-evaluation}
Pedro Rodriguez, Joe Barrow, Alexander~Miserlis Hoyle, John~P. Lalor, Robin
  Jain, and Jordan Boyd-Graber.
\newblock Evaluation examples are not equally informative: How should that
  change {NLP} leaderboards?
\newblock In {\em Proceedings of the 59th Annual Meeting of the Association for
  Computational Linguistics (Volume 1: Long Papers)}, pages 4489--4504, Online,
  August 2021. Association for Computational Linguistics.

\bibitem[\protect\citeauthoryear{{\c{S}}ahin}{2022}]{csahin2022augment}
G{\"o}zde~G{\"u}l {\c{S}}ahin.
\newblock To augment or not to augment? a comparative study on text
  augmentation techniques for low-resource nlp.
\newblock {\em Computational Linguistics}, 48(1):5--42, 2022.

\bibitem[\protect\citeauthoryear{Sambasivan \bgroup \em et al.\egroup
  }{2021}]{sambasivan2021everyone}
Nithya Sambasivan, Shivani Kapania, Hannah Highfill, Diana Akrong, Praveen
  Paritosh, and Lora~M Aroyo.
\newblock Everyone wants to do the model work, not the data work: Data cascades
  in high-stakes ai.
\newblock In {\em proceedings of the 2021 CHI Conference on Human Factors in
  Computing Systems}, pages 1--15, 2021.

\bibitem[\protect\citeauthoryear{Siminyu \bgroup \em et al.\egroup
  }{2020}]{siminyu2020ai4d}
Kathleen Siminyu, Sackey Freshia, Jade Abbott, and Vukosi Marivate.
\newblock Ai4d--african language dataset challenge.
\newblock {\em arXiv preprint arXiv:2007.11865}, 2020.

\bibitem[\protect\citeauthoryear{Siminyu \bgroup \em et al.\egroup
  }{2021}]{siminyu2021ai4d}
Kathleen Siminyu, Godson Kalipe, Davor Orlic, Jade Abbott, Vukosi Marivate,
  Sackey Freshia, Prateek Sibal, Bhanu Neupane, David~I Adelani, Amelia Taylor,
  et~al.
\newblock Ai4d--african language program.
\newblock {\em arXiv preprint arXiv:2104.02516}, 2021.

\bibitem[\protect\citeauthoryear{Singh \bgroup \em et al.\egroup
  }{2024a}]{l2024indicgenbench}
Harman Singh, Nitish Gupta, Shikhar Bharadwaj, Dinesh Tewari, and Partha
  Talukdar.
\newblock Indicgenbench: A multilingual benchmark to evaluate generation
  capabilities of llms on indic languages.
\newblock In {\em Proceedings of the 62nd Annual Meeting of the Association for
  Computational Linguistics (Volume 1: Long Papers)}, pages 11047--11073, 2024.

\bibitem[\protect\citeauthoryear{Singh \bgroup \em et al.\egroup
  }{2024b}]{singh2024aya}
Shivalika Singh, Freddie Vargus, Daniel D’souza, B{\"o}rje~F Karlsson,
  Abinaya Mahendiran, Wei-Yin Ko, Herumb Shandilya, Jay Patel, Deividas
  Mataciunas, Laura O’Mahony, et~al.
\newblock Aya dataset: An open-access collection for multilingual instruction
  tuning.
\newblock In {\em Proceedings of the 62nd Annual Meeting of the Association for
  Computational Linguistics (Volume 1: Long Papers)}, pages 11521--11567, 2024.

\bibitem[\protect\citeauthoryear{Sloane \bgroup \em et al.\egroup
  }{2022}]{sloane2022participation}
Mona Sloane, Emanuel Moss, Olaitan Awomolo, and Laura Forlano.
\newblock Participation is not a design fix for machine learning.
\newblock In {\em Proceedings of the 2nd ACM Conference on Equity and Access in
  Algorithms, Mechanisms, and Optimization}, pages 1--6, 2022.

\bibitem[\protect\citeauthoryear{Snyder}{2019}]{snyder2019literature}
Hannah Snyder.
\newblock Literature review as a research methodology: An overview and
  guidelines.
\newblock {\em Journal of business research}, 104:333--339, 2019.

\bibitem[\protect\citeauthoryear{Susanto \bgroup \em et al.\egroup
  }{2025}]{susanto2025sea}
Yosephine Susanto, Adithya~Venkatadri Hulagadri, Jann~Railey Montalan,
  Jian~Gang Ngui, Xianbin Yong, Wei~Qi Leong, Hamsawardhini Rengarajan, Peerat
  Limkonchotiwat, Yifan Mai, and William~Chandra Tjhi.
\newblock Sea-helm: Southeast asian holistic evaluation of language models.
\newblock In {\em Findings of the Association for Computational Linguistics:
  ACL 2025}, pages 12308--12336, 2025.

\bibitem[\protect\citeauthoryear{Taiuru}{2021}]{taiuru2021kaitiakitanga}
Karaitiana Taiuru.
\newblock Kaitiakitanga m{\=a}ori data sovereignty licences, 2021.

\bibitem[\protect\citeauthoryear{Thu \bgroup \em et al.\egroup
  }{2016}]{thu2016introducing}
Ye~Kyaw Thu, Win~Pa Pa, Masao Utiyama, Andrew Finch, and Eiichiro Sumita.
\newblock Introducing the asian language treebank (alt).
\newblock In {\em Proceedings of the Tenth International Conference on Language
  Resources and Evaluation (LREC'16)}, pages 1574--1578, 2016.

\bibitem[\protect\citeauthoryear{Yu \bgroup \em et al.\egroup
  }{2026}]{yu2026afriquellm}
Hao Yu, Tianyi Xu, Michael~A Hedderich, Wassim Hamidouche, Syed~Waqas Zamir,
  and David~Ifeoluwa Adelani.
\newblock Afriquellm: How data mixing and model architecture impact continued
  pre-training for african languages.
\newblock {\em arXiv preprint arXiv:2601.06395}, 2026.

\bibitem[\protect\citeauthoryear{Zheng \bgroup \em et al.\egroup
  }{2023}]{zheng2023judging}
Lianmin Zheng, Wei-Lin Chiang, Ying Sheng, Siyuan Zhuang, Zhanghao Wu, Yonghao
  Zhuang, Zi~Lin, Zhuohan Li, Dacheng Li, Eric~P. Xing, Hao Zhang, Joseph~E.
  Gonzalez, and Ion Stoica.
\newblock Judging {LLM}-as-a-judge with {MT}-bench and chatbot arena.
\newblock In {\em Advances in Neural Information Processing Systems},
  volume~36, 2023.

\end{thebibliography}
 
\end{document}